\documentclass{ceurart}
\usepackage{listings}
\usepackage{tcolorbox}

\begin{document}
\copyrightyear{2025}
\copyrightclause{Copyright for this paper by its authors.
  Use permitted under Creative Commons License Attribution 4.0
  International (CC BY 4.0).}
\conference{CLEF 2025 Working Notes, 9 -- 12 September 2025, Madrid, Spain}

\title{TeamCMU at Touch\'e: Adversarial Co-Evolution for Advertisement Integration and Detection in Conversational Search}

\title[mode=sub]{Notebook for the Touch\'e Lab at CLEF 2025}

\author{To Eun Kim}[%
orcid=0000-0002-2807-1623,
email=toeunk@cs.cmu.edu,
]
\address{Carnegie Mellon University, Pittsburgh, PA, USA}

\author{João Coelho}[%
orcid=0009-0001-6207-1934,
email=jmcoelho@cs.cmu.edu,
]
\fnmark[1]

\author{Gbemileke Onilude}[%
orcid=0009-0006-7264-1693,
email=gonilude@cs.cmu.edu,
]
\fnmark[1]

\author{Jai Singh}[%
orcid=0009-0007-6276-3029,
email=jsingh2@andrew.cmu.edu,
]
\fnmark[1]

\fntext[1]{Equal contribution.}

\newcommand{\tekcomment}[1]{\textcolor{red}{[TEK: #1]}}
\newcommand{\jccomment}[1]{\textcolor{blue}{[JC: #1]}}
\newcommand{\gocomment}[1]{\textcolor{green}{[GO: #1]}}
\newcommand{\jscomment}[1]{\textcolor{teal}{[JS: #1]}}

\newcommand{\touche}{Touché\xspace}

\newcommand{\qaSystem}{\texttt{QA System}\xspace}
\newcommand{\adRewriter}{\texttt{Ad-Rewriter}\xspace}
\newcommand{\adClassifier}{\texttt{Ad-Classifier}\xspace}

\newcommand{\naiveSynthetic}{\textit{NaiveSynthetic}\xspace}
\newcommand{\structuredSynthetic}{\textit{StructuredSynthetic}\xspace}

%




\newcommand{\query}{q}
\newcommand{\collection}{\mathcal{D}}

\newcommand{\rModel}{\mathcal{R}}
\newcommand{\doc}{d}
\newcommand{\retrievalResults}{z}

\newcommand{\promptGenerator}{\phi_p}
\newcommand{\augmentedPrompt}{\overline{\query}}
\newcommand{\qaLLMModel}{\mathcal{F}}
\newcommand{\qaLLMResponse}{r}

\newcommand{\adItem}{a}

\newcommand{\adRewriterModel}{\mathcal{G}}
\newcommand{\rewriterResponse}{y}

\newcommand{\adClassifierModel}{\mathcal{H}}
\newcommand{\classifierScore}{s}

\newcommand{\relevanceMetric}{\mu_r}
\newcommand{\utilityMetric}{\mu_u}

\begin{abstract}
As conversational search engines increasingly adopt generation-based paradigms powered by Large Language Models (LLMs) and Retrieval-Augmented Generation (RAG), the integration of advertisements into generated responses presents both commercial opportunities and challenges for user experience. Unlike traditional search, where advertisements are clearly delineated, generative systems blur the boundary between informational content and promotional material, raising concerns around transparency and trust. In this work, we propose a modular pipeline for advertisement management in RAG-based conversational systems, consisting of an ad-rewriter for seamless ad integration and a robust ad-classifier for detection. We leverage synthetic data to train high-performing classifiers, which are then used to guide two complementary ad-integration strategies: supervised fine-tuning of the ad-rewriter and a best-of-N sampling approach that selects the least detectable ad-integrated response among multiple candidates. Our evaluation focuses on two core questions: the effectiveness of ad classifiers in detecting diverse ad integration strategies, and the training methods that best support coherent, minimally intrusive ad insertion. Experimental results show that our ad-classifier, trained on synthetic advertisement data inspired by marketing strategies and enhanced through curriculum learning, achieves robust detection performance. Additionally, we demonstrate that classifier-guided optimization, through both fine-tuning and best-of-N sampling, significantly improves ad stealth, enabling more seamless integration. These findings contribute an adversarial co-evolution framework for developing more sophisticated ad-aware generative search systems and robust ad classifiers.
\end{abstract}

\begin{keywords}
Conversational Search,
Retrieval-Augmented Generation,
LLM,
Advertisement,
Classification
\end{keywords}

\maketitle

\section{Introduction}

Conversational search engines powered by Large Language Models (LLMs) \cite{radlinski2017theoretical} and Retrieval-Augmented Generation (RAG) \cite{lewis2020rag, kim2024reml} are increasingly integrating advertisements into responses to enhance monetization. As these systems shift toward generation-driven paradigms, the inclusion of advertising content in LLM outputs has become both a timely and underexplored area, especially as state-of-the-art industry systems move toward ad-supported deployments \cite{perplexity2024advertising, openai2025shopping}. Given that advertising has historically served as the primary revenue stream for search engines \cite{gleason2024search}, this transition raises critical questions about how to embed ads in generated content without compromising response utility or user trust. Unlike traditional search interfaces, where sponsored content is explicitly demarcated, generative systems risk blurring the line between organic information and promotional material, potentially obfuscating ad presence in the absence of clear markers \cite{Zelch24adinGenIR}.

Despite its significance for the future of commercial LLM systems, advertisement integration and transparency in LLM-generated responses remain insufficiently studied. While prior work has introduced auction frameworks for generative ads and investigated methods for detecting LLM-generated advertisements \cite{Dubey24kddAuction, Schmidt24DetectingAds}, comprehensive generation-side strategies remain limited. In addition, foundational insights from marketing research, such as the distinctions between explicit vs. implicit advertising and soft vs. hard selling \cite{yi1990direct, shapiro2001memory, okazaki2010measuring}, have yet to be meaningfully incorporated into generative model design. 
It also remains unclear whether existing ad-detection systems \cite{post2015comparative, shiller2018effect}, originally developed for traditional media, can generalize to the diverse and increasingly subtle forms of ad integrated in LLM-generated contents. Furthermore, recent efforts that rely on naive ad insertion strategies \cite{Schmidt24DetectingAds} may risk compromising response quality and user experience.

\begin{figure*}[t!]
\centering
\includegraphics[width=\linewidth]{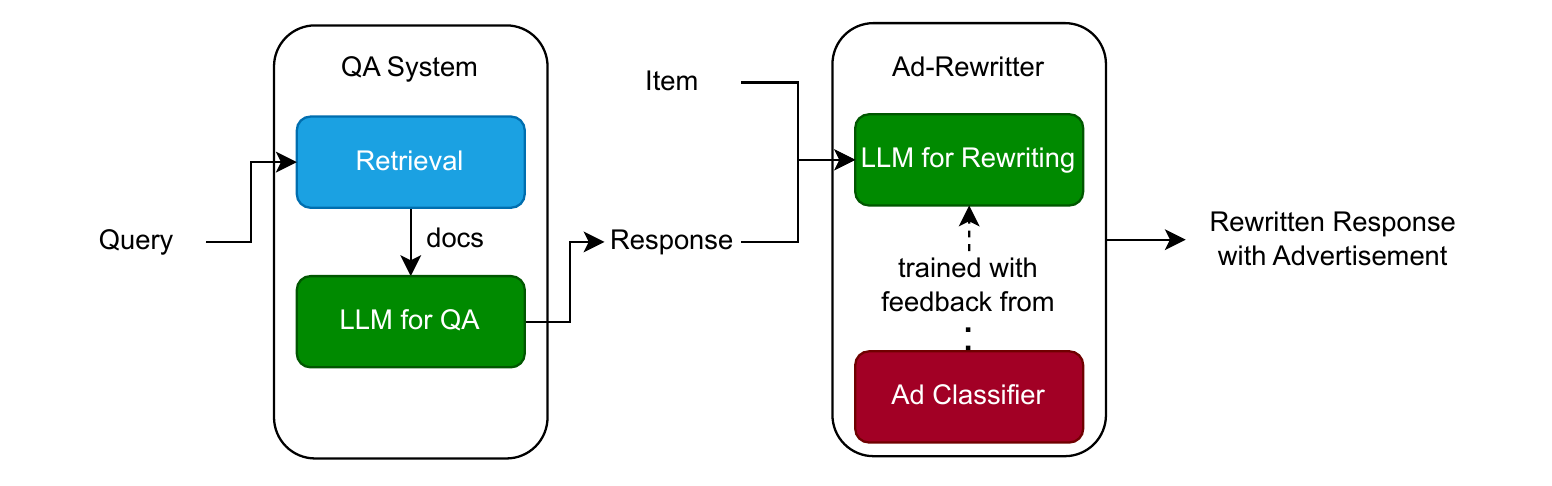}
\caption{Overall Pipeline. 
A user query is first processed by the \qaSystem to generate a base response. When an item is specified for advertisement, it is passed along with the base response to the \adRewriter, which produces an ad-integrated version of the response. To further reduce ad detectability, we apply a best-of-$N$ sampling strategy, selecting the rewritten response with the lowest ad probability as predicted by the \adClassifier.
}
\label{fig:pipeline}
\end{figure*}

To address these challenges, we participate in both sub-tasks (generation and classification) of the \textit{Advertisement in Retrieval-Augmented Generation} shared task at the \touche lab \cite{kiesel2025toucheoverview}, CLEF 2025, where our systems were submitted via the TIRA platform \cite{froebe2023tira}.
We propose a modular pipeline (Figure \ref{fig:pipeline}) for advertisement management in RAG-based conversational systems. Our architecture consists of a standalone RAG-based \qaSystem, followed by an \adRewriter that integrates advertisements into the generated responses, and an \adClassifier trained to detect them. 

The \adRewriter is experimented in three variants: a zero-shot version prompted directly for ad integration, a supervised fine-tuning (SFT) variant trained with feedback from a robust ad-classifier, and a zero-shot version enhanced with best-of-N sampling, where the final response is selected from multiple candidates based on the classifier's ad probability scores.

In this \textit{adversarial co-evolution} setup, the robustness of the \adClassifier is critical to the effectiveness of the \adRewriter. To enhance the robustness of a classifier, we augment the provided dataset with carefully curated synthetic data, including hard positive and hard negative instances. The enhanced classifier is then used as feedback mechanisms to guide the optimization of the \adRewriter across different implementation strategies.

Our study explores the feasibility of this framework through two central research questions:
\begin{itemize}
\item \textbf{RQ1}: How can we train an \adClassifier that achieves robust classification performance across diverse types of ad-integrated responses?
\item \textbf{RQ2}: How can we develop an \adRewriter that enables seamless ad integration while minimizing the likelihood of ad detection?
\end{itemize}

Through our experiments, we demonstrate the effectiveness of training the \adClassifier using hard positive and hard negative synthetic data to improve robustness. We also show that training the \adRewriter in an adversarial setup (\textit{i.e.,} using feedback from the robust classifier) leads to more effective and less detectable ad integration.
We publicly release our code for further research.\footnote{\url{https://github.com/kimdanny/TeamCMU-AdRAG}}

\section{Related Work}

In this section, we survey related work on open domain QA and emerging strategies for advertising in LLM-based search applications.

\subsection{Open Domain QA}
Early QA systems relied on extracting answer spans from retrieved documents and machine reading comprehension models \cite{seo2017bidirectional}. Models such as DrQA \cite{chen2017drqa} set the foundation for modern QA by improving retrieval and answer span prediction.
Large language models (LLMs) revolutionized QA by enabling zero-shot and few-shot learning \cite{brown2020language}. While these models provide high-quality answers, challenges remain in evaluation, due to large but semantically sound answers that differ from the gold label~\citep{DBLP:conf/acl/KamallooDCR23}, and other LLM-related problems such as hallucination~\citep{DBLP:conf/emnlp/LiCZNW23}.
Retrieval-Augmented Generation (RAG)~\citep{lewis2020rag}, a specialized method for generation as part of a retrieval-enhanced machine learning strategy~\citep{kim2024reml, zamani2022reml, diaz2024remltutorial} combines retrieval mechanisms with LLMs to improve factual correctness and response relevance, especially in knowledge-intensive task such as QA and fact-checking~\citep{petroni-etal-2021-kilt}.

The MS-MARCO dataset~\citep{bajaj2016msmarco} has driven significant web QA advancements, with state-of-the-art methods employing dense retrieval~\citep{karpukhin-etal-2020-dpr, DBLP:conf/iclr/XiongXLTLBAO21} and contrastive learning to optimize both response quality and retrieval accuracy. Recent hybrid architectures that strategically combine LLMs with dense retrievers have demonstrated measurable improvements over standalone GPT-3.5 or LLaMA-7B prompting~\citep{li2024unigen}. Similar approaches~\citep{DBLP:journals/tacl/WangHJ024, yin2025arrquestionansweringlarge} have also obtained state-of-the-art results on other benchmarks~\citep{DBLP:conf/acl/JoshiCWZ17,kwiatkowski-etal-2019-natural}, showing the versatility of retrieval-enhanced language models across domains.

\subsection{Advertisement in the Era of LLMs}
Search systems are actively employing LLMs to display search results~\citep{sun2023chatgptsearch}. In the era of LLMs, revenue generation through online advertising within LLM-generated response is gaining attention. In response, researchers are starting to investigate auctions and advertising strategies in the context of LLM-based search systems.
\citet{Dubey24kddAuction} studied an auction framework ensuring higher bidders receive greater ad placement in LLM outputs.
Inspired by them, \citet{hajiaghayi2024adauction} examined advertisement auctions with a focus on RAG, by considering both relevance (from the retriever) and bids when allocating ads within generated responses.
\citet{soumalias2024aggregationAds} proposed an auction framework where advertisers influence LLM responses through reinforcement learning from human feedback.
The detection of generated ad content is also a growing research area. \citet{Schmidt24DetectingAds} introduced the Webis Generated Native Ads 2024 dataset, focusing on identifying LLM-generated ads.

\section{Problem Definition}

In this section, we restate the two sub-tasks for the \textit{Advertisement in Retrieval-Augmented Generation} shared task at the \touche lab \cite{kiesel2025toucheoverview} with more details.

\subsection{Sub-task 1: Ad-Augmented QA}
In Sub-task 1, the QA system is provided with an open-domain web query, a set of relevant passages, and a set of external items to be advertised. The objective is to build a system that leverages the passages to answer the query while incorporating an advertisement for one of the provided items.
If $m$ items are given, the system should generate $m$ independent answers, each integrating a distinct item. These advertisements should be seamlessly woven into the response and difficult to detect as ads. Additionally, the system must be capable of generating standard, non-advertising answers when no items are provided, ensuring those responses do not exhibit ad-like characteristics.

\paragraph{\touche-25 Advertisement-in-Retrieval-Augmented-Generation (Ad-RAG) Dataset}\label{data:ad_rag}

The Ad-RAG dataset comprises approximately 3,000 queries, for which systems are required to generate both ad-augmented and standard responses.\footnote{\url{https://zenodo.org/records/14699130}} The queries are typically short phrases that describe a topic or product (\textit{e.g.,} \textit{``good triceps workout equipment"}, \textit{``corvette z06"}). For half of the queries, no advertisements are needed, but just an informative response. For the remaining queries, each requires the inclusion of, on average, two advertisements. Information about the items to be advertised is provided in the form of short descriptions averaging six words. Each query is supported by up to 100 passages retrieved from the MS MARCO v2.1 dataset~\citep{bajaj2016msmarco} using BM25 retrieval~\citep{DBLP:journals/ftir/RobertsonZ09}.
These queries in the Ad-RAG dataset were obtained from the Webis-Ads dataset~\citep{Schmidt24DetectingAds}, which will be used in Sub-task 2.

\subsection{Sub-task 2: Ad Detection}
The objective of Sub-task 2 is to determine whether a given response contains an embedded advertisement. Specifically, the system receives a response as input and performs binary classification to predict whether the response includes an advertisement or is purely informative. An effective classifier should be robust to subtle ad insertions, ensuring that even seamlessly integrated advertisements can be accurately detected.

\paragraph{Webis Generated Native Ads 2024 (Webis-Ads) Dataset}\label{sec:webis}
The Webis-Ads dataset \citep{Schmidt24DetectingAds} was created to train an ad-blocker system for conversational search engines.
This dataset comprises approximately 7,500 queries, along with responses generated by Microsoft Copilot and YouChat. For half of these queries, a second version of the response was produced by prompting GPT-4 to insert advertisements without altering the original informative content. As a result, the dataset includes 7,500 responses without ads and 3,800 responses with ads.

Notably, the data in this dataset is relatively easy to fit. In our preliminary experiments, a simple DeBERTa-based text classifier \cite{he2021debertav3} achieved around 98\% accuracy on held-out data, suggesting that the naive ad-insertion strategy used to construct the dataset results in easily detectable patterns. This observation motivates the need for more challenging training data. To address this, we construct synthetic hard positive and hard negative examples, which we discuss in detail in the following sections.

\section{Methodology}

In this section, we describe our methodology for building a more robust \adClassifier and leveraging it as a feedback mechanism to improve the effectiveness of the \adRewriter.

\subsection{Pipeline Overview}
Figure~\ref{fig:pipeline} presents an overview of our system. Given a user query $\query$, the retrieval-augmented \qaSystem retrieves relevant passages and generates an initial response $\qaLLMResponse$ without any advertisements.\footnote{In the \touche competition, retrieved documents are provided. As a result, we do not evaluate retrieval effectiveness and simply use the top-$k$ passages.} When a specific item $\adItem$ is provided for advertisement, the \adRewriter module $\adRewriterModel$ modifies the base response $\qaLLMResponse$ to seamlessly incorporate the promotional content, yielding a rewritten response $\rewriterResponse$.

The \adRewriter can operate in several modes: it can be 1) prompted to produce a rewritten response directly, 2) guided by best-of-N sampling \cite{stiennon20learningtosummarize, nakano2021webgpt} using feedback from a trained \adClassifier, or 3) fine-tuned through supervised learning using training data generated with classifier feedback.

\subsection{QA System}

The \qaSystem is responsible for generating contextually relevant responses to open-domain queries prior to any advertisement integration. While a typical QA pipeline involves both retrieval and generation, in the \touche competition setting, retrieved passages are provided. Thus, we directly proceed to the generation step using the top-$k$ passages.

Given the top-$k$ retrieved passages $\retrievalResults$, we prompt a language model $\qaLLMModel$ to synthesize a coherent and self-contained response.\footnote{Qwen2.5-7B-Instruct is used as a language model in our experiment.} Prompts are constructed using a prompt generation function $\promptGenerator^{\qaLLMModel}(\query, \retrievalResults)$, which is designed to elicit cohesive and informative responses from the model. 
The base QA output, $\qaLLMResponse$, serves as the input to the \adRewriter module when advertisement integration is required.
Prompt used for the response generation can be found in Appendix \ref{sec:app:qasystem}.

\subsection{Ad-Classifier}
The \adClassifier $\adClassifierModel$ is formulated as a standard binary text classification task: given a query $\query$ and its corresponding response $\qaLLMResponse$, the model predicts whether the response contains an advertisement. To build increasingly robust classifiers, we incrementally expand the training data with progressively harder examples derived from multiple sources.

The initial version (\textbf{V0.0}) was trained solely on the Webis-Ads dataset \citep{Schmidt24DetectingAds}; a simple DeBERTa-based classifier \cite{he2021debertav3} achieved strong performance on held-out data.\footnote{V0.0: \url{https://huggingface.co/jmvcoelho/ad-classifier-v0.0}}
However, we found that it failed to generalize to more naturally embedded or implicit forms of advertising.

To address this limitation, we introduced two complementary types of synthetic training data. The first, \naiveSynthetic dataset, involves prompting an LLM to insert fictional advertisements into baseline QA responses without constraints, resulting in a wide variety of superficially embedded ads. With this data, we trained two classifiers: \textbf{V0.1} and \textbf{V0.2}. 

The second, \structuredSynthetic dataset, incorporates real-world product entities sourced from Wikipedia. Drawing on advertising and marketing literature~\cite{yi1990direct, shapiro2001memory, okazaki2010measuring}, we extract descriptive features and generate two categories of training examples: (i) hard positives, where the product is promoted through indirect or implicit language, and (ii) hard negatives, which are neutral informative passages about the product with no advertising intent.
With the \structuredSynthetic dataset, we train successive versions of the classifier using combinations of the Webis-Ads, \naiveSynthetic, and \structuredSynthetic datasets: \textbf{V0.3}, \textbf{V0.4}, and \textbf{V0.5}. In the last two versions (V0.4, V0.5), we incorporate curriculum learning \cite{bengio2009curriculum} based on classification difficulty, as estimated by the output logits from an earlier classifier (V0.1). This training strategy produces classifiers with improved generalization and robustness to diverse ad integration strategies, including those grounded in effective marketing practices.

\subsubsection{Creation of NaiveSynthetic Data}
\naiveSynthetic data generation follows the original Webis-Ads dataset approach, \textit{i.e.}, given an answer without an advertisement, prompt an LLM to inject an ad. The query generation prompts include no specific item; rather, the LLM is instructed to generate an advertisement of an item that fits the context, which may result in the creation of fictional products.
To promote diversity, we use a combination of 5 different LLMs: GPT-4o, Gemma-2-9B-it\footnote{\url{https://huggingface.co/google/gemma-2-9b-it}}, LLaMA-3.1-8B-Instruct\footnote{\url{https://huggingface.co/meta-llama/Llama-3.1-8B-Instruct}}, Qwen2.5-7B-Instruct\footnote{\url{https://huggingface.co/Qwen/Qwen2.5-7B-Instruct}}, and Mistral-7B-Instruct.\footnote{\url{https://huggingface.co/mistralai/Mistral-7B-Instruct-v0.1}}
Moreover, we devise 12 different prompts for ad insertion, targeting various advertising strategies (\textit{e.g.,} direct, indirect, explicit, implicit, hard-sell and soft-sell). An example prompt for \naiveSynthetic query generation can be found in Appendix \ref{sec:appendix:navive-synth}.

Using this setup, we trained two versions of the classifier. Both V0.1 and V0.2 leverage the same set of LLMs. However, V0.1 uses a single prompt for data generation, while V0.2 randomly samples from the full pool of 12 prompts. The HuggingFace model pages contain the prompts used for insertion.\footnote{V0.1: \url{https://huggingface.co/jmvcoelho/ad-classifier-v0.1}; V0.2: \url{https://huggingface.co/jmvcoelho/ad-classifier-v0.2}}

\subsubsection{Creation of StructuredSynthetic Data}
We generate \structuredSynthetic dataset through the following steps:

\begin{enumerate}
    \item \textit{Systematic collection of product entities from Wikipedia.}

    We manually select Wikipedia "infobox" namespaces likely to contain product-related pages that can be advertised (\textit{e.g.}, `product', `brand', `camera', `automobile'), collecting a total of 25 infoboxes.

    To ensure that each page within these namespaces refers to a real product (e.g., \textit{iPhone}) rather than a general concept (e.g., \textit{Mobile phone}), we filter pages using Wikidata properties that strongly indicate "product-ness" (e.g., P162 – producer, P593 – model number). This allows us to curate a set of non-fictional product entities along with their associated Wikipedia content.

    For each verified entity, we retrieve its release year, rank the entities by recency, and retain only those released in or after the year 2000.\\

    \item \textit{Wikipedia article summarization and extraction of key promotional features.}

    For each selected entity, we prompt a GPT-4o model to summarize the corresponding Wikipedia page and extract key features and qualities suitable for promotional purposes.\\

    \item \textit{Creation of hard positives (indirect and implicit advertisements) and hard negatives (factual, non-promotional texts).}

    Drawing on insights from advertising literature, we generate two types of data using GPT-4o:  
    \begin{itemize}
        \item Hard positives: Indirect and implicit advertisements.  
        \item Hard negatives: Factual and informative descriptions without promotional intent.
    \end{itemize}
    List of infoboxes, Wikidata properties, and the prompts used for hard positive and negative query generations can be found in Appendix \ref{sec:appendix:structuredSynthetic}.
\end{enumerate}

Using this setup, we trained three versions of the classifier. In V0.3, the classifier was trained on a combined dataset consisting of Webis-Ads, \naiveSynthetic, and \structuredSynthetic instances. In V0.4, we applied curriculum learning \cite{bengio2009curriculum}, where instance difficulty was determined by the V0.1 model. Finally, V0.5 used the same training regime as V0.4, but balanced the \naiveSynthetic and \structuredSynthetic instances by upsampling the \structuredSynthetic dataset. Further details are available on the corresponding HuggingFace model pages.\footnote{V0.3: \url{https://huggingface.co/teknology/ad-classifier-v0.3}; V0.4: \url{https://huggingface.co/teknology/ad-classifier-v0.4}; V0.5: \url{https://huggingface.co/teknology/ad-classifier-v0.5}}

\subsection{Ad-Rewriter}\label{subsec:ad-rewriter}

The \adRewriter module $\adRewriterModel$ takes as input a query $\query$, an ad-free QA response $\qaLLMResponse$, and a product or service to be advertised $\adItem$. These elements are combined into a prompt, denoted as $\promptGenerator^{\mathcal{G}}(\query, \qaLLMResponse, \adItem)$, which conditions the rewriting process. The goal of the \adRewriter is to produce a fluent, contextually relevant, and minimally intrusive ad-integrated version of the original response.

\paragraph{Method 1: Zero-shot rewriting}
Our initial implementation relies on a prompt-based zero-shot rewriting, exploring advertisement strategies from the marketing literature, such as direct vs. indirect and explicit vs. implicit advertising. Prompt used for the rewriting can be found in Appendix \ref{sec:app:adrewriter-prompt}.

\paragraph{Method 2: Supervised fine-tuning-based rewriting}
To move beyond prompt engineering, we construct a training dataset using our synthetic query generation pipeline. For each $(\query, \qaLLMResponse, \adItem)$ triplet, we generate five candidate ad-integrated responses: $\rewriterResponse_i \sim \adRewriterModel(\promptGenerator^{\mathcal{G}}(\query, \qaLLMResponse, \adItem))$ for $i \in {1..5}$, where $\adRewriterModel$ can be various LLMs with different temperature. Each rewritten response $\rewriterResponse_i$ is then scored by the ad-classifier $\adClassifierModel$, which estimates the likelihood that the response contains an advertisement: $\adClassifierModel(\rewriterResponse_i)$.

We adopt a supervised fine-tuning (SFT) regime in which the objective is to train the model to prefer completions with lower predicted ad probability. Formally, we define the optimal response $y^*$ and the negative log-likelihood loss $\mathcal{L}_{NLL}$ as:
\begin{equation}
    y^* = \text{argmin}_{y \in \{y_1,\ldots,y_5\}} \adClassifierModel(y)
\end{equation}
\begin{equation}
    \mathcal{L}_{NLL} = - \log \text{P}( y^* \;|\;  \promptGenerator^{\mathcal{G}}(\query, \qaLLMResponse, \adItem)) .
\end{equation}

\paragraph{Method 3: Zero-shot rewriting with Best-of-N sampling}
Due to the computational cost of fine-tuning a language model, we apply a best-of-N sampling strategy in the zero-shot method. We set our generation temperature above zero, and the model produces a diverse set of $N$ candidate rewrites for each input. Each candidate response is then evaluated using a trained ad-classifier, which assigns an ad probability score. We select the response with the lowest predicted ad probability as the final rewritten output. In our experiments, we use $N = 10$.

This feedback loop, where classifiers guide the training of rewriters, forms the backbone of our approach, which aims to result in more natural ad integration within the generated responses.

\section{Experiments}

\subsection{Experimental Setup}\label{sec:eval}

For Sub-task 1, systems are evaluated adversarially using a hidden classifier maintained by the task organizers. The primary evaluation metric is advertisement detection accuracy, where a lower score indicates more successful ad integration, \textit{i.e.}, the advertisement is more difficult to detect. As the official classifier is unavailable during development, we employ a series of in-house \adClassifier (V0.0–V0.5) as proxy evaluators for iterative tuning and model comparison. Specifically, we compare detection accuracy across the three different \adRewriter strategies (zero-shot, zero-shot with BoN, and supervised fine-tuning-based rewriting) to assess the detectability of inserted ads. This setup also serves to evaluate the performance of the \adClassifier for Sub-task 2 using standard binary classification accuracy.
For \adClassifier, we use pre-trained DeBERTa model \cite{he2021debertav3}, and for \qaSystem and \adRewriter, we use Qwen2.5-7B-Instruct \cite{qwen2}.

\subsection{Results}

\paragraph{RQ1: How can we train an \adClassifier that achieves robust classification performance across diverse types of ad-integrated responses?}

Before proceeding to ad rewriting, we first identify which classifiers perform well across a range of ad-integration strategies. To evaluate this, we test six versions of classifiers on responses generated using three ad rewriting approaches: a pure zero-shot method, a fine-tuned rewriter, and a zero-shot method with best-of-N (BoN) sampling. Each approach is evaluated under two different generation temperature settings. Recall that these rewriters modify the base QA response using various advertising techniques (\textit{e.g.,} indirect promotion through storytelling), producing a diverse set of ad-integrated outputs.

\begin{table*}[ht]
\centering
\caption{Ad-detection accuracy (\%) across different versions of classifiers (V0.0–V0.5) under three ad rewriting methods with varying generation temperatures. This table highlights both the robustness of the classifiers and the effectiveness of the ad rewriting strategies. Among the classifiers, versions V0.1 and V0.4 demonstrate strong performance across a diverse set of ad-integrated responses. Regarding ad-rewriting methods, SFT and Zero-Shot-BoN-based \adRewriter show low ad detection accuracy, indicating more seamless integration compared to the zero-shot approach across all classifier versions.
}
\label{tab:main-results}
\small
\begin{tabular}{cccccccc}
\toprule
\textbf{Rewriting Method} & \textbf{Temperature} & \textbf{V0.0} & \textbf{V0.1} & \textbf{V0.2} & \textbf{V0.3} & \textbf{V0.4} & \textbf{V0.5} \\
\midrule
\multirow{2}{*}{Zero-Shot} & 1.0 & 34.93 & 86.77 & 73.98 & 85.31 & 91.85 & 69.36 \\
                                     & 0.5 & 33.33 & 83.43 & 67.87 & 80.78 & 89.02 & 76.03 \\
\midrule
\multirow{2}{*}{SFT} & 1.0 & 18.87 & 66.16 & 51.73 & 61.32 & 61.94 & 49.79 \\
                                              & 0.5 & 16.99 & 55.22 & 41.55 & 51.34 & 46.24 & 38.54 \\
\midrule
\multirow{2}{*}{Zero-Shot-BoN} & 1.0 & 32.24 & 54.35 & 41.90 & 52.34 & 49.78 & 46.93 \\
                                              & 0.5 & 30.98 & 51.57 & 40.14 & 50.24 & 47.68 & 45.34 \\
\bottomrule
\end{tabular}
\end{table*}

Table~\ref{tab:main-results} presents the ad classification accuracy for each classifier version across the different rewriting approaches.
When comparing V0.0, which is trained without synthetic data or specialized training methods such as curriculum learning, to the other versions, we observe a substantial improvement in ad detection accuracy. This indicates that synthetic data inspired by marketing strategies can effectively enhance classifier performance. In particular, versions V0.1 and V0.4 achieve consistently high accuracy across both temperature settings, underscoring the value of combining diverse synthetic training data with curriculum learning.

\paragraph{RQ2: How can we develop an \adRewriter that enables seamless ad integration while minimizing the likelihood of ad detection?}
In addition to the basic zero-shot ad rewriter, we develop two enhanced versions: the SFT and Zero-Shot-BoN-based rewriter. Both approaches rely on a classifier to provide feedback, either during inference in the BoN setting or during training data construction for the SFT approach. Based on the classification performance observed on outputs from the pure Zero-Shot rewriter, we select V0.4 as the most robust classifier and use it both for scoring candidate responses in the best-of-N sampling strategy and for fine-tuning the rewriter.

As shown in Table~\ref{tab:main-results}, both the SFT and Zero-Shot-BoN-based rewriters lead to a substantial reduction in ad detection accuracy across all classifier versions when compared to the basic Zero-Shot rewriter. These results demonstrate that incorporating classifier feedback, either at inference time or during training, is effective in achieving more seamless ad integration. In fact, these methods are able to deceive robust classifiers in approximately 50 percent of cases. A paired Wilcoxon signed-rank test \cite{woolson2005wilcoxon} confirms that this reduction is statistically significant for all classifiers, indicating that classifier-guided generation can reliably produce responses that are more difficult to detect as advertisements.

\paragraph{Shared Task Results}
For Sub-Task 1 (ad generation), the primary evaluation metric designated by the lab organizers is recall; lower recall indicates more subtle ad insertion, and thus a better-performing system. In addition to automatic metrics, the organizers manually reviewed 100 responses per submission to validate classifier labels and ensure the legitimacy of each run.
We submitted our system based on the Zero-Shot-BoN approach, using the Qwen2.5-7B-Instruct model \cite{qwen2} for both the \qaSystem and the \adRewriter, and employing the V0.4 classifier for best-of-N sampling. Out of six submissions, including a baseline, our system ranked 3rd by recall (see Table~\ref{tab:subtask1-results}). If F1-Score had been the ranking criterion, our system would have placed 2nd.

For Sub-Task 2 (ad classification), the primary evaluation metric set by the lab organizers is F1-Score.
We submitted our V0.4 classifier for evaluation. Among 16 runs, including baselines, our classifier ranked 3rd in terms of F1-Score (see Table~\ref{tab:subtask2-results}).

\begin{table*}[ht]
\centering
\caption{Sub-Task 1 (ad generation) evaluation results by Touch\'e.}
\begin{tabular}{l l c c c}
\toprule
\textbf{Team} & \textbf{Run} & \textbf{Precision} & \textbf{Recall} & \textbf{F1-Score} \\
\midrule
JU-NLP & ORPO\_Mistral7b\_v2 & 1.000 & 0.721 & 0.838 \\
JU-NLP & ORPO\_Mistral7b & 0.995 & 0.830 & 0.905 \\
\textbf{TeamCMU} & Adrewriting-BestOfN & 0.821 & 0.858 & 0.839 \\
Git Gud & Qwen2.5 7B V2 & 0.960 & 0.910 & 0.935 \\
Git Gud & Qwen3 4B V2 & 0.984 & 0.918 & 0.950 \\
Baselines & generate-baseline & 0.796 & 0.996 & 0.885 \\
\bottomrule
\end{tabular}
\label{tab:subtask1-results}
\end{table*}
\begin{table*}[ht]
\centering
\caption{Sub-Task 2 (ad detection) evaluation results by Touch\'e.}
\begin{tabular}{l l c c c}
\toprule
\textbf{Team} & \textbf{Run} & \textbf{Precision} & \textbf{Recall} & \textbf{F1-Score} \\
\midrule
JU-NLP & DebertaFineTuned & 0.788 & 0.758 & 0.773 \\
Git Gud & Deberta-Large-V2 & 0.983 & 0.473 & 0.639 \\
\textbf{TeamCMU} & deberta-synthetic-curriculum & 0.945 & 0.479 & 0.636 \\
Git Gud & Roberta-Large & 0.985 & 0.460 & 0.627 \\
Baseline & minilm-baseline & 0.728 & 0.482 & 0.580 \\
Pirate Passau & MPnet-finetuned & 0.399 & 0.917 & 0.556 \\
Pirate Passau & Tf-IDF-Logestic-Regression & 0.395 & 0.734 & 0.514 \\
JU-NLP & Finetuned\_MPNET\_v2 & 0.977 & 0.346 & 0.511 \\
JU-NLP & Finetuned\_MPNET & 0.305 & 1.000 & 0.467 \\
Baseline & naive-bayes-10 & 0.307 & 0.968 & 0.467 \\
Baseline & naive-bayes-25 & 0.319 & 0.638 & 0.425 \\
Pirate Passau & All-mini-LM-v2-finetuned & 0.664 & 0.294 & 0.408 \\
Git Gud & Deberta Large & 0.312 & 0.355 & 0.332 \\
Baseline & naive-bayes-40 & 0.367 & 0.257 & 0.302 \\
Pirate Passau & all-mini+Random-forest & 0.341 & 0.022 & 0.042 \\
Pirate Passau & LLM-llama3.1 & 0.500 & 0.000 & 0.001 \\
\bottomrule
\end{tabular}
\label{tab:subtask2-results}
\end{table*}

\section{Discussion}
\paragraph{Methodological Implications}
Our findings across the two research questions and the strong evaluation results highlight the effectiveness of using carefully curated synthetic data and classifier-guided training in managing ad integration within generative systems. For ad detection, we show that classifiers trained on synthetic data inspired by marketing strategies---particularly when combined with curriculum learning---achieve significantly more robust performance across a diverse set of ad-integration styles. On the generation side, incorporating classifier feedback, either through best-of-N sampling or during supervised fine-tuning, leads to ad-integrated responses that are substantially harder to detect. These results suggest that adversarial training dynamics between rewriters and classifiers can be effective in shaping both components for more seamless and harder-to-detect ad insertion.

Among these generation strategies, we also observe that responses generated at lower temperatures tend to yield lower ad detection rates. One possible explanation for this pattern is that the model produces more coherent and well-structured responses at lower temperatures \cite{brown2020language}, allowing ad insertions to blend more naturally with the surrounding content. In contrast, higher temperatures introduce greater variability, which can result in phrasing or transitions that are less contextually aligned, making the presence of advertisements more noticeable to the classifier.

\paragraph{Limitations} A key constraint of this study is the reliance on synthetic data generated by LLMs necessitates more rigorous validation and incorporation of more challenging scenarios to ensure robustness. The binary nature of the current advertisement classifier may also fall short in fully capturing nuanced or context-dependent advertisements. 
Additionally, the metric of ad detectability is grounded in classifier performance. However, human users may perceive ads differently, and what evades a model may still be obvious to a human reader.

\paragraph{Ethical Considerations} 
This work reveals that advertisements can be seamlessly integrated into LLM-generated responses in ways that are difficult even for strong classifiers to detect. While this demonstrates the technical feasibility of subtle ad insertion, it also underscores the importance of accompanying such capabilities with appropriate transparency controls. Without explicit labeling or disclosure mechanisms, users may be unknowingly exposed to persuasive content, potentially diminishing trust in conversational systems \cite{Zelch24adinGenIR}. Moreover, false positives from ad classifiers risk misclassifying informative content, which could disadvantage legitimate content providers. Ethical challenges are amplified when ads appear in sensitive contexts, such as mental health or emergency-related queries, or when cultural stereotypes and provider-side exposure imbalances propagate through system components. These findings highlight the need for careful design choices and deployment safeguards to ensure that stealthy ad integration does not come at the cost of user agency or marketplace fairness \cite{mehrotra2018towards}.

\paragraph{Future Direction} Future work can address current limitations through comprehensive validation of synthetic data using approaches like system rank correlation and linguistic analysis \cite{he2025tip}. Beyond any technical improvements, future implementations can explore more realistic scenarios involving retrieval based on dynamic ad bidding information \cite{hajiaghayi2024adauction}. Moreover, evaluating and ensuring provider-side fairness will be essential for maintaining a balanced and sustainable advertisement ecosystem, demanding rigorous assessment of both provider-consumer dynamics and systemic biases \cite{kim2025fairragimpactfair}.
\section{Conclusion}
We show that fine-tuning an advertisement classifier using synthetic query data inspired by marketing strategies, along with progressively harder detection examples, significantly enhances its robustness and effectiveness in identifying seamlessly integrated ads. Notably, we find that feedback from such a well-trained classifier, whether used during test-time sampling or as part of the training objective, can be leveraged to guide ad generators that strategically evade detection, successfully deceiving even strong classifiers. This adversarial dynamic underscores both the potential and the challenge of developing reliable and transparent advertisement in LLM-based search systems.

\begin{acknowledgments}
  We thank Professor Eric Nyberg, Professor Teruko Mitamura, and Kimihiro Hasegawa for their valuable feedback during the development of our system.
\end{acknowledgments}

\section*{Declaration on Generative AI}
  %
 %
 During the preparation of this work, the authors used generative AI in order to identify and correct grammatical errors and typos. The authors reviewed and edited the content as needed and take full responsibility for the publication's content. 

\bibliography{99-ref}

\appendix

\section{Computational Cost}\label{sec:appendix:computation}
All inference experiments, except for ablations on models larger than 7B parameters, were conducted on a single L40 GPU.
For efficiency, we distributed experiments across 2 to 32 GPUs depending on availability. Training was performed using 8 L40 GPUs for supervised fine-tuning on 7B parameter models, while ad-classifier training required only a single GPU.

\section{Prompt for \qaSystem response generation}\label{sec:app:qasystem}
\begin{tcolorbox}
Answer the following web query, given the context.\\
Context: \{context\}.\\
Query: \{query\}.\\

Reply only with a "well formed answer", i.e., human-like and coheseve. The answer should be written in a style where advertisements are easy to be integrated. 
Do not include any advertisements or explicit advertisement space, just write with the stylistic choice in mind.
\end{tcolorbox}

\section{Prompt for \adRewriter rewritten response generation}\label{sec:app:adrewriter-prompt}
\begin{tcolorbox}
Consider this existing response to a conversational search query, and modify it to include an implicit advertisement.\\
The advertisement should:\\
- Mention this \{type\}: \{item\};\\
- Highlight these qualities: \{qualities\};\\
- Blend naturally with the original response;\\
- Not appear overly promotional;\\
- Be hard to detect by current textual ad-detection systems.\\

Original query: \{query\}\\
Original response: \{response\}\\
Return only the modified response with the advertisement integrated.
\end{tcolorbox}

\newpage
\section{Details on Synthetic Data Generation}
\subsection{Naive Synthetic Data}\label{sec:appendix:navive-synth}

Example prompt for \naiveSynthetic query generation:
\small\begin{tcolorbox}
Task: You are tasked with subtly integrating an advertisement into a search query response. The goal is to make the advertisement feel natural and helpful within the context of the response, not disruptive or overtly promotional.
First, you should define the item to advertise. You should keep in mind the context of the query and original response. Consider the following advertisement qualities when choosing the product:\\
* Product/Service Category.\\
* Specific Advertisement Theme.\\
* Key Selling Points/Qualities to Emphasize.\\

Instructions for Integration:\\
1.  Contextual Relevance is Key:  The advertisement must feel relevant to the search query and the existing response text.  Think about how the advertised product/service genuinely relates to the user's needs implied by the query.\\
2.  Subtle and Natural Language:  Use language that blends seamlessly with the original response. Avoid overly promotional or sales-oriented phrasing. Focus on informative and helpful language.\\
3.  Focus on Benefits, Not Just Features: Instead of directly listing qualities, rephrase them as benefits the user might gain in relation to the search query and response context.\\
4.  Strategic Placement: Choose the most natural and impactful location(s) within the response to subtly introduce the advertisement. This might involve:\\
    * Briefly modifying an existing sentence to subtly incorporate the advertisement.\\
    * Adding a short, relevant phrase or clause to an existing sentence.\\
    * In rare cases, adding a very short, contextual sentence (only if it feels truly natural).\\
5.  Maintain Original Meaning:  Ensure the core meaning and factual accuracy of the original response remain unchanged. The advertisement should enhance, not distort, the original information.\\
6.  Review for Subtlety: Before returning the response, critically evaluate if the advertisement feels genuinely subtle and integrated. If it feels forced or obvious, refine your approach.\\

Output: Return **only** the modified response with the subtly integrated advertisement.\\

Search Query: \{query\}\\
Original Response: \{response\}\\
Modified Response:
\end{tcolorbox}

The following versions of \adClassifier were trained using the \naiveSynthetic data:
\begin{itemize}
    \item \textbf{V0.1}: 
    \begin{itemize}
        \item Multiple models, single data generation prompt.
        \item \url{https://huggingface.co/jmvcoelho/ad-classifier-v0.1}
    \end{itemize}
    
    \item \textbf{V0.2}: 
    \begin{itemize}
        \item Multiple models, multiple data generation prompt.
        \item \url{https://huggingface.co/jmvcoelho/ad-classifier-v0.2}
    \end{itemize}
\end{itemize}

The HuggingFace model pages contain the prompts used for insertion.

\subsection{Structured Synthetic Data}\label{sec:appendix:structuredSynthetic}

\subsubsection{List of infoboxes selected}
product, brand, automobile, motorcycle, tractor, calculator, computing device, keyboard, software, camera, mobile phone, night vision device, synthesizer, tool, watch, pinball, toy, film, book, Asian comic series, comic, musical, furniture, video game, drug.

\subsubsection{List of Wikidata properties selected}
P50 (author),  
P86 (composer),  
P110 (illustrator),  
P123 (publisher),  
P162 (producer),  
P170 (creator),  
P176 (manufacturer),  
P178 (developer),  
P179 (product series),  
P287 (designed by),  
P593 (model number),  
P676 (lyricist),  
P943 (programmer),  
P3640 (National Drug Code),  
P4087 (MyAnimeList manga ID),  
P8731 (AniList manga ID),  
P9618 (AlternativeTo software ID),  
P9897 (App Store age rating), and   
P12969 (game designer)  

\subsubsection{Hard positive creation prompt}
\begin{tcolorbox}
Your task is to generate an indirect and implicit advertisement for a \{infobox\_name\} named \{product\_name\}.

The advertisement\\
    * must not indicate that it is an advertisement or promotional content.\\
    * must include the \{infobox\_name\} name, \{product\_name\}.\\
    * must avoid any direct call to action.\\
    * must be brief and contained within one paragraph.\\
    * may present the \{infobox\_name\} as part of natural, conversational, or informational content, or as a synthetic personal experience that could occur in real life.\\
    * may use testimonial or storytelling styles that describe the experiences of people with \{page\_title\}.\\
    * may include detailed, scientific/research-backed statements.\\
    
The following information about \{page\_title\} may be useful for your writing:\\
\{summary\}\\
The advertisement can implicitly promote some of the following aspects of \{page\_title\}: \{key\_features\}\\
Write only the advertisement without any explanations.
\end{tcolorbox}

\subsubsection{Hard negative creation prompt}
\begin{tcolorbox}
Your task is to write a concise, informative text about a \{infobox\_name\} named \{product\_name\}.

The text:\\
    * must focus on delivering factual information.\\
    * must not include expressions of preference or favoritism toward \{page\_title\} and should focus solely on the facts.\\
    * must include the name \{product\_name\} at least once.\\
    * can mention other \{infobox\_name\}s related to \{page\_title\} to provide comprehensive information about the subject.\\

The following information about \{page\_title\} may be useful for your writing:\\
\{summary\}\\
Write only the informative text without any explanations.
\end{tcolorbox}

The following versions of \adClassifier were trained with the \structuredSynthetic data:
\begin{itemize}
    \item \textbf{V0.3}: 
    \begin{itemize}
        \item \url{https://huggingface.co/teknology/ad-classifier-v0.3}
    \end{itemize}

    \item \textbf{V0.4}: 
    \begin{itemize}
        \item Trained by curriculum learning.
        \item \url{https://huggingface.co/teknology/ad-classifier-v0.4}
    \end{itemize}
    
    \item \textbf{V0.5}: 
    \begin{itemize}
        \item Trained by curriculum learning and data balancing.
        \item \url{https://huggingface.co/teknology/ad-classifier-v0.5}
    \end{itemize}
\end{itemize}

\end{document}